# SafeStep: AI-powered Travel Assistance for Elderly People with Frailty or Dementia

Azul Debenedetti,[1] David Gamez,[1] Franco Such[2] and Nik Kairinos[2]

1. Department of Computer Science, Middlesex University, London, UK
2. Fountech AI Limited, Evripidou 12A, Agia Zoni, 3031 Limassol, Cyprus

**Abstract.** More than a million people in the UK suffer from frailty or dementia, which severely compromise their ability to travel in urban environments. This paper presents SafeStep, an AI-driven travel system that assists elderly users with their journeys. At the core of SafeStep is a novel travel graph representation, which integrates route planning with predictive modelling. For each stage of a journey, the system (i) generates personalized failure scenarios using a combination of LLMs and the Anticip8 behavioral prediction engine, (ii) proposes targeted interventions, and (iii) estimates the impact of interventions on outcome probabilities. This enables SafeStep to select interventions that maximize the likelihood of the person reaching their destination. SafeStep was evaluated through experiments on travel graph generation and a field study involving 26 real-world journeys. Results showed that combining Anticip8 for failure prediction with GPT-based models for intervention evaluation yields the most reliable performance. User feedback indicated that SafeStep improves confidence and perceived safety during travel, although interface usability needs to be improved for the target demographic. In the future, we would like to improve and release SafeStep. The AI system that was developed for SafeStep could be applied in other areas, such as mental health, career coaching and addiction treatment.

## 1 Introduction

In the UK, approximately 1.3 million elderly people are affected by frailty, and 982,000 are living with dementia. People living with frailty are at increased risk of fall injuries, which reduces their ability to travel. People with dementia can easily get overwhelmed and lost on public transport systems, where they struggle to follow previously familiar routes. These issues make it more likely that people with frailty and dementia will stay at home, which reduces their physical activity and has a negative impact on their mental health.

This paper describes SafeStep, a tool that is designed to assist people with frailty and dementia on their journeys. The SafeStep AI system plans the journey using Google Maps, and then executes the following steps for each stage of the journey:

1. Predict how the journey stage could go wrong for a person with frailty or dementia.
2. Generate interventions that could prevent these failure scenarios.
3. Test the interventions to identify the ones that are most likely to prevent failure scenarios.

This planning is done using a combination of a large language model (GPT-5) and a commercial action prediction system (Anticip8). The result is a travel graph for the journey, which is used by the app to guide the user and deploy interventions that help keep their journey on track. Different methods for generating the travel graph were explored and the app was tested in a live setting by a small number of volunteers. The results indicate that an improved version of SafeStep could have a significant impact on the travel problems that are encountered by people with frailty and dementia.

The first part of the paper gives background information about the issues faced by people with frailty and dementia and covers previous work on the prediction of human actions. Next, the travel graph generation is described in detail, followed by a brief overview of the app. Sections 5 and 6 cover experiments on travel graph generation and field testing, and the paper concludes with a discussion of the current limitations of SafeStep, and suggestions about future work.

# 2 Background

## 2.1 Frailty and Dementia

The United Kingdom's population was estimated to be 69.3 million in 2024, with 19% of people belonging to the 65+ age group [1]. According to the British Geriatrics Society, frailty is a common condition that affects over 10% of the 65+ demographic and over 50% of the 85+ demographic [2]. The characteristics of frailty include progressive loss of muscle strength and lifestyle and health changes caused by stressors such as falling, infections, and confusion. People living with frailty often struggle with poorly maintained pavements and a lack of support on public transport. Frail people also develop a fear of falling that makes them less inclined to leave home [3].

According to the UK Alzheimer’s Society, in 2024 there were an estimated 982,000 people living with dementia in the UK [4]. The most common forms of dementia are vascular dementia (damage to the brain’s blood vessels), and Alzheimer’s disease (abnormal buildup of proteins in the brain). Both types progressively lead to memory loss, planning difficulties, personality changes and disorientation. When travelling on public transport, dementia patients often struggle to follow familiar routes and wrongly estimate the passing of time. Requesting help and buying tickets can also be harder for dementia patients. According to the Alzheimer's Society, the anxiety and panic caused by stressful situations, such as getting lost, can make symptoms of dementia worsen [5]. Getting lost or its assumed risk is a leading cause of permanent care home residency for dementia patients [6].

### 2.2 Travel Assistance for People with Frailty or Dementia

Tools like Google Maps have revolutionized urban mobility for the general population. However, they are not designed for people with frailty and dementia. Mistakes, such as taking a wrong turn or early disembarkation, are communicated to the user after they happen, which makes it difficult for the user to get back on track. The interfaces of these applications presuppose a calm rational user and fail to account for how easily overwhelmed a person with dementia can be. LLM-based solutions to these problems have been proposed, including a website that uses LLMs to provide travel assistance [7] and the Mobility-LLM travel assistant for the elderly [8]. A review of LLM-based chatbots designed to assist the elderly is given by Tahir et al. [9].

### 2.3 Prediction of Human Actions

A substantial amount of research has been done on the prediction of human actions in video data, which is targeted at applications like security surveillance and autonomous driving. For example, Kong and Fu [10] review previous work on action recognition in human figures and Kolekar et al. [11] summarize research on predicting the actions of people driving vehicles. Work has also been done on predicting human actions inside the home [12], human decision making [13] and crime [14].

Large language models (LLMs) are trained with large quantities of text that describe human actions. If enough context is provided, LLMs can make plausible predictions about what people will do next. However, they can struggle to understand the physical world, which places important constraints on human actions. To address this issue Takeyama et al. [15] built a system that combined LLM-based action prediction with trajectory data, which significantly improved prediction performance. Related work was done by Graule and Isler [16], who enhanced LLM-based action prediction with a semantic map of the environment. LLMs have also been used to augment action prediction in video sequences [17], and research has been done on accident prediction in the elderly using images and a multimodal LLM [18]. A graph-based approach, similar to the one described in this paper, was developed by Song et al. [19], who constructed a graph from smart device logs and used a LLM to add detail to this graph to better understand users' actions and intentions.

Researchers have also fine-tuned LLMs to improve their ability to predict human actions. For example, Kolluri et al [20] fine-tuned a model using 2.9 million responses from 400,491 participants in 210 open-source social science experiments. The resulting model was able to produce predictions that were 26% more aligned with the distributions of human responses. A commercial product in this space has been developed by Fountech, whose Anticip8 (https://anticip8.ai/) predictive AI system is designed to forecast human behavior and decision-making. The inputs to Anticip8 are a profile of the individual, recent activity and factors that could impact their behavior. Anticip8 uses this data to predict the next actions of the individual or to sort actions according to their probability.

# 3 Travel Graph Generation

At the heart of SafeStep is an AI system that plans the user's journey, identifies failure scenarios, and suggests interventions that could prevent the failure scenarios. This information is represented as a travel graph, which is used by the app (see Section 4) to assist the person on their journey. An example travel graph is given in Fig. 1.

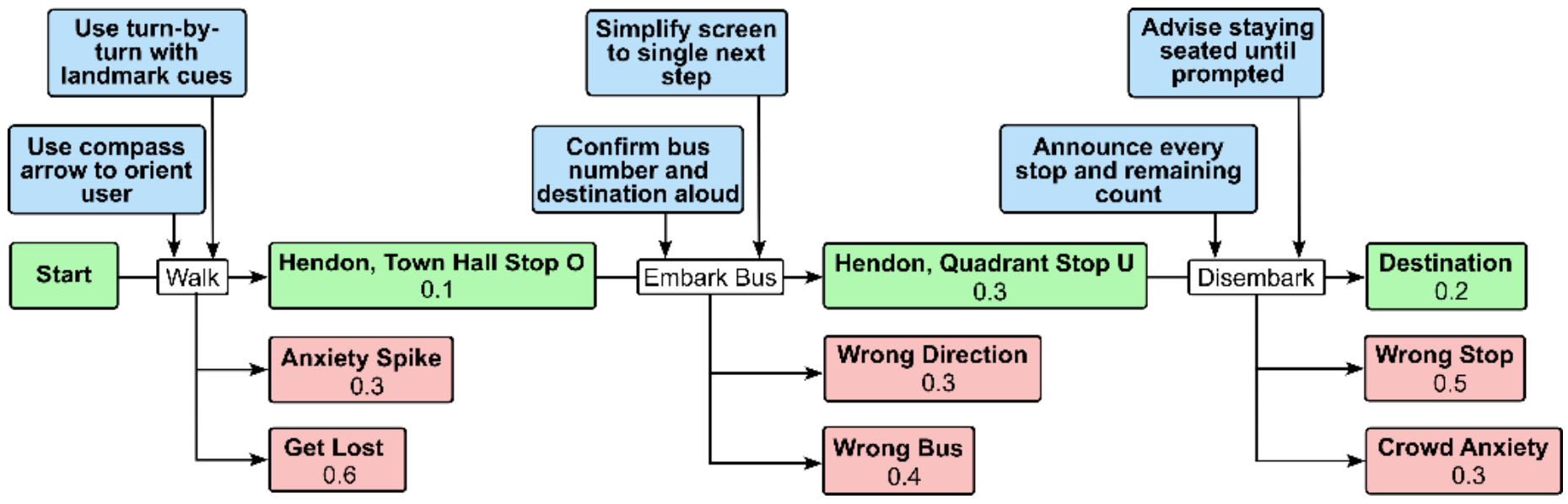

**Fig. 1.** Example of travel graph for person with dementia.[1] The green nodes are the route generated by Google Maps. Red nodes are failure scenarios, where the journey could go wrong, and blue nodes are interventions that could prevent the failure scenarios. The number on each node is the estimated probability that it will occur.

## 3.1 Journey Plan

The first step in the generation of the travel graph is a call to Google Maps to get a journey plan. In London, UK, where the system was tested, this typically consists of walking and a combination of public transport options (bus, train, tube, etc.).

## 3.2 Failure Scenarios

The next step predicts some of ways in which each stage of the journey could go wrong. For example, a person with dementia might miss the bus or get on the wrong bus; a person with frailty might slip on the pavement. These nodes are generated by prompting Anticip8 or GPT-5 with the user's profile (examples given in Table 2) and requesting failure scenarios for each stage of the journey. The models are also requested to estimate the probability of each failure scenario.

## 3.3 Interventions

For each journey stage, GPT-5 was prompted to generate interventions that could prevent the failure scenarios. Anticip8 was not used for this part of the graph generation process because it is designed to predict actions from a given context, not actions that could change the context.

[1] The graphs in Fig. 1 and Fig. 2 are manually constructed for illustrative purposes. Real examples of SafeStep travel graphs are given in [21].

### 3.4 Impact of Interventions

The final step uses Anticip8 or GPT-5 to predict the impact of interventions on the user's journey. Anticip8 or GPT-5 are prompted with the user's profile and each intervention in turn, and the models are requested to produce revised probabilities of the failure nodes. This enables the system to select the intervention that is most likely to keep the user's journey on track. This process is illustrated in Fig. 2.

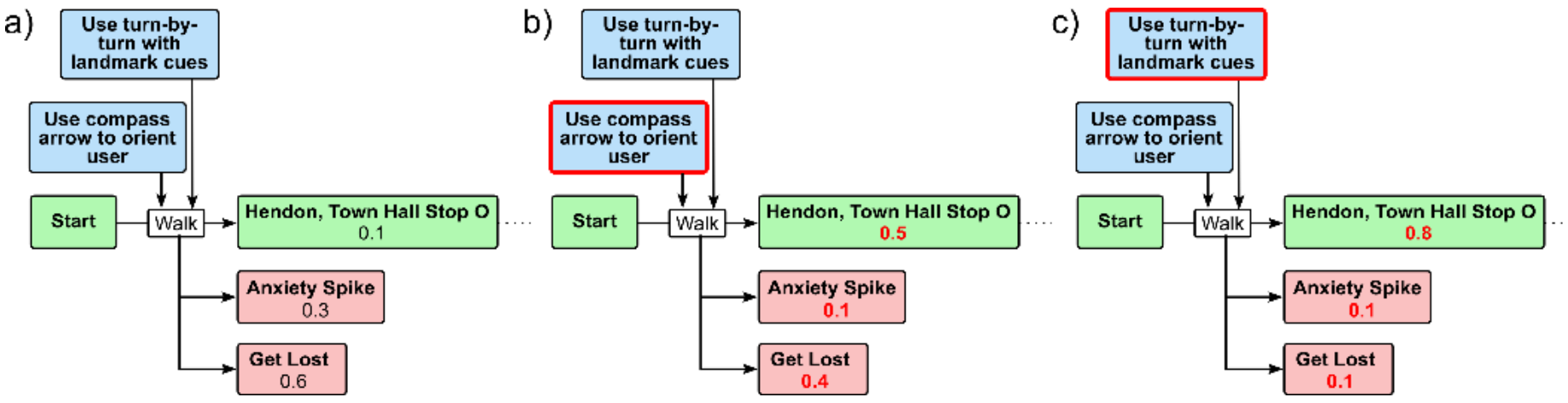

**Fig. 2.** Impact of interventions. Each intervention is selected in turn and added to the GPT-5 or Anticip8 prompt. The intervention that is predicted to maximize the chances of success is chosen for the user. a) First stage of the Fig. 1 travel graph showing estimated probabilities without interventions. b) The compass intervention is added to the prompt, and the AI models are asked to estimate the probabilities of the success and failure nodes, given this intervention. The compass reduces anxiety and increases the probability that the user will reach the bus stop. c) Next, the turn-by-turn navigation with landmark cues intervention is tried. This reduces anxiety and is selected for the user because it has the best chance of preventing them from getting lost.

### 3.5 Graph Visualization Tool

To help develop the system, a visualization tool was built that enables users to generate and visualize travel graphs for different user profiles and journeys. In the current prototype the user's health condition (frailty or dementia) is hard coded as two test user profiles (see Table 2). The graph visualization tool is shown in Fig. 3.

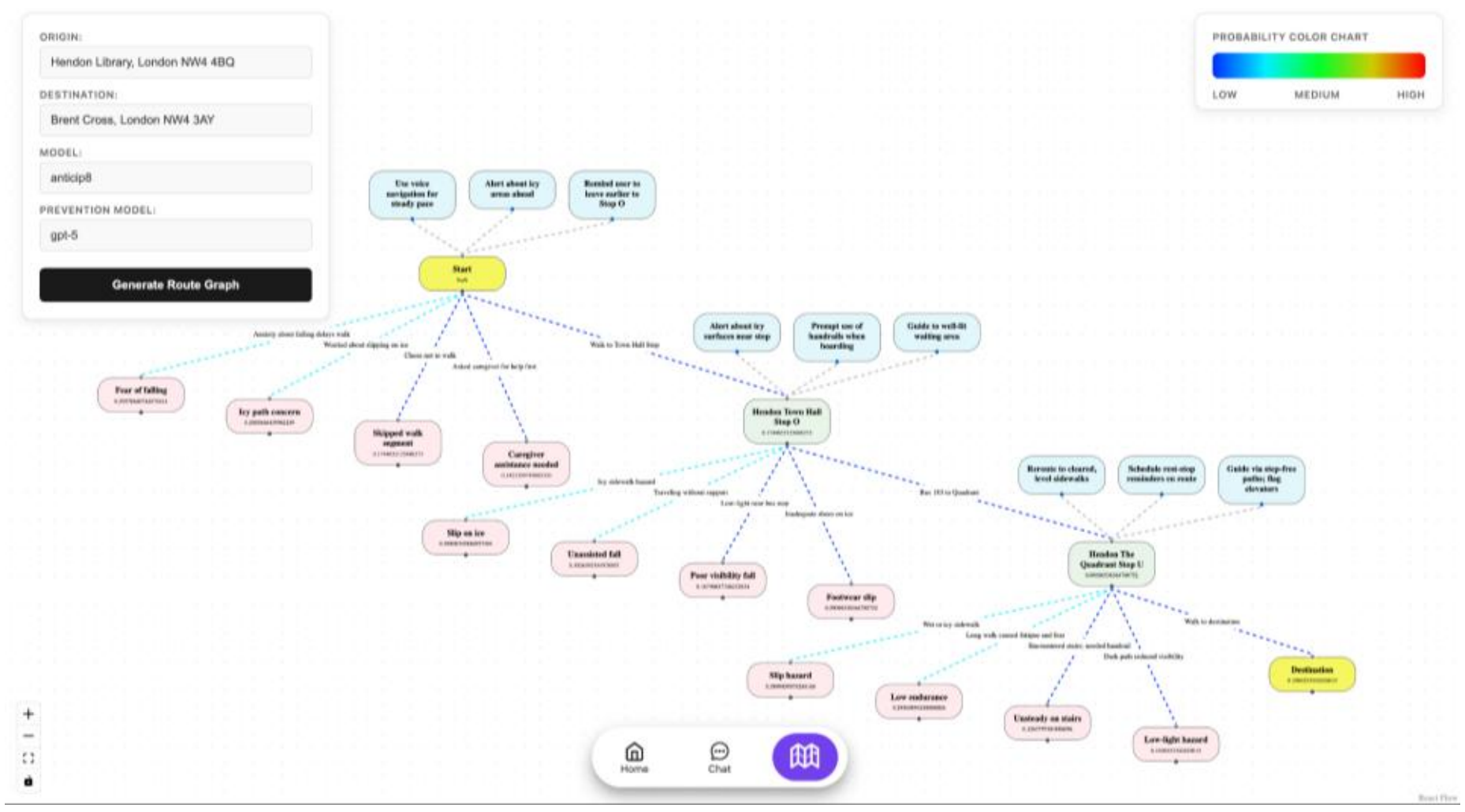

**Fig. 3.** Travel graph visualization tool.

## 4 SafeStep App

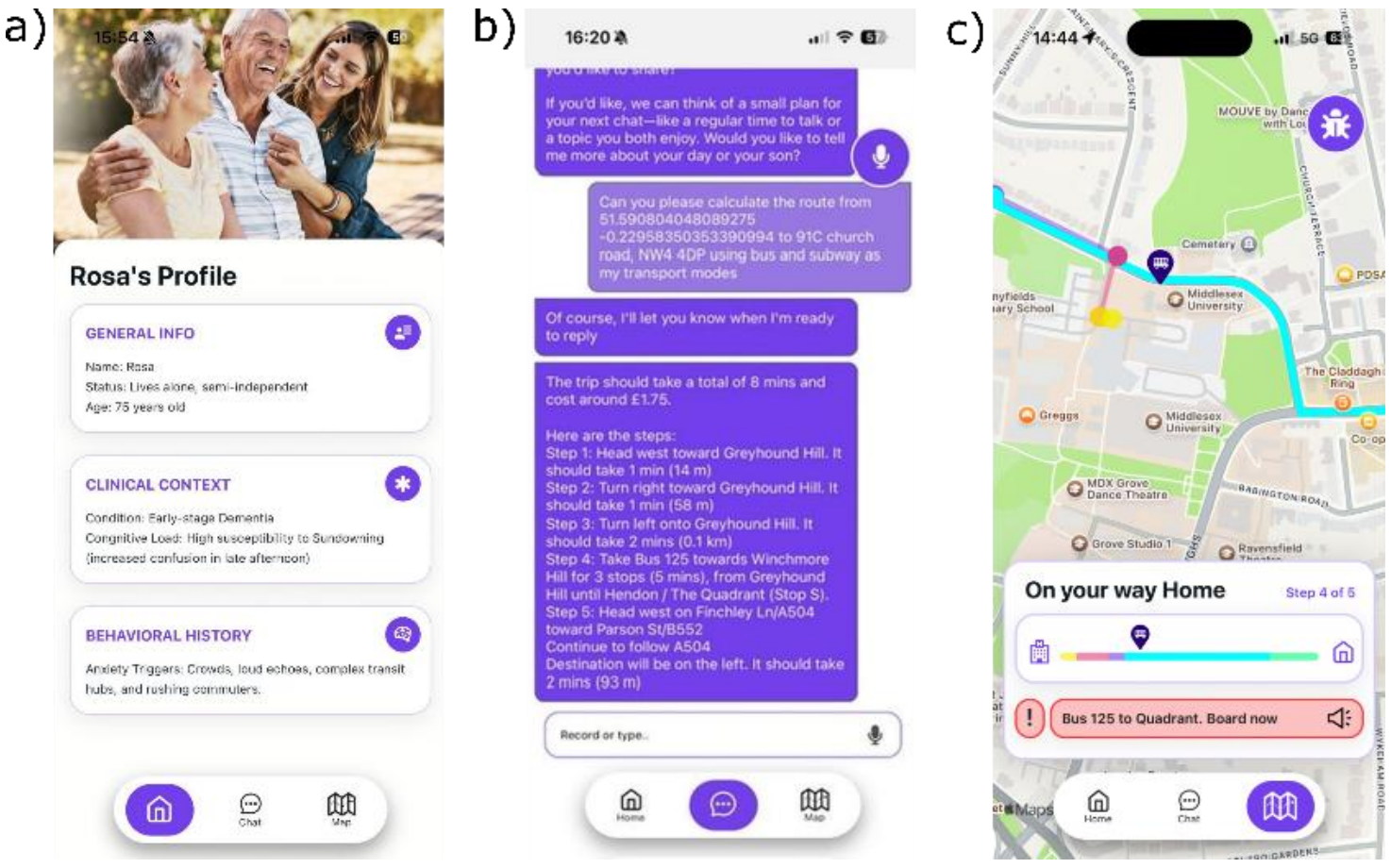

**Fig. 4.** SafeStep app. a) Profile information. b) Text- and audio-based journey planner. c) Map showing journey status and interventions.

Users interact with SafeStep through the app shown in Fig. 4. Users enter their health information on the profile page. After they have entered their journey details, the app requests the travel graph from the server and uses it to guide the person on their journey, applying the best intervention at each stage. The journey is tracked on a simple map interface. The app also has calendar integration that enables routes to appointments to be pre-calculated.

# 5 Travel Graph Generation Experiments

## 5.1 Methodology

Experiments were carried out to discover the best way of generating failure nodes and predicting intervention effects. For each profile given in Table 2, different combinations of Anticip8 and GPT-5 were used to generate the failure nodes and predict how the probabilities would change when the interventions were applied. This was done for the three journeys described in Table 1. The experiments were run using the graph visualization tool.

**Table 1.** Journeys used in Travel Graph Experiments

| Route | Origin | Destination |
|---|---|---|
| 1 | Trolley Park, Brent Cross, London NW2 6GJ | 51.5860469, -0.2071016 |
| 2 | Hendon Library, London NW4 4BQ | 91C Church Road, NW4 4DP |
| 3 | The Regent's Park, London | Manor House, Friern Barnet Ln, London N20 0NL |

**Table 2.** Profiles for travel graph generation tests

| Data | Dementia Profile | Frailty Profile |
|---|---|---|
| Name | Rosa | Harold |
| Age | 75 | 82 |
| Living status | Lives alone, semi-independent | Lives alone, receives part-time caregiver support |
| Medical condition | Early-stage dementia | Physical frailty |
| Physiological problem | High susceptibility to "sundowning" (increased confusion in late afternoon) | Low physiological reserve, high fatigue after minimal exertion |
| Symptoms | • Short-term memory lapses<br>• Spatial disorientation<br>• Executive function decline<br>• Sensory overstimulation | • Reduced muscle strength<br>• Slow gait<br>• Poor balance and fall risk<br>• Low endurance<br>• Delayed recovery after illness |
| Mobility limitations | • Bus Accuracy: 20% failure rate<br>• Directional Logic: 25% failure rate<br>• Waypoint Memory: 10% failure rate | • Needs rest after 5-10 mins walking<br>• Requires handrail on stairs (40% difficulty without)<br>• Uses cane outdoors |
| Anxiety triggers | Crowds, loud echoes, complex transit hubs, and rushing commuters. | Fear of falling, icy sidewalks, long standing times, rushed environments, and poorly lit spaces. |

### 5.2 Results

The results for different combinations of Anticip8 and GPT-5 for the dementia and frailty profiles are discussed in Table 3 and Table 4. Example graphs for the different combinations are given in [21]. For both profiles the best results were obtained when Anticip8 generated the failure nodes and GPT-5 assessed the impact of interventions on the probabilities of the failure nodes. Anticip8's poor performance on the probability predictions is unsurprising because it is designed to generate and rank actions, rather than predict the probabilities of actions. The calibration of probabilistic outputs is an active area of Anticip8 development.

**Table 3.** Results for frailty profile

| Failure Node Generation | Probability Calculation | Results |
|---|---|---|
| GPT-5 | GPT-5 | Failure nodes tended to be highly generic, with standard risks, such as "trip on curb". Probabilities could be over-optimistic, reaching 1.0 likelihood of success in some cases. |
| Anticip8 | Anticip8 | Anticip8 successfully integrated the user's clinical profile in the generation process, resulting in highly personalized failure nodes. The probability recalculations based on interventions were inconsistent and success probabilities often decreased when applying preventions, even when they were highly relevant to both failure nodes and profiles. |
| GPT-5 | Anticip8 | This combination had the worst attributes of each model. GPT-5 generated generic failure nodes that missed key aspects of the user's profile. Anticip8's prediction of the interventions' effects was arbitrary and inconsistent. |
| Anticip8 | GPT-5 | This combination successfully balanced clinical personalization and better probability weighting. By employing Anticip8 for generation, the system captured user-specific risks, while GPT-5 made plausible predictions about the effects of interventions on the probabilities. |

**Table 4.** Results for dementia profile

| Failure Node Generation | Probability Prediction | Results |
|---|---|---|
| GPT-5 | GPT-5 | This combination resulted in fast superficial results. GPT-5 generates generic cognitive risks that fail to take into consideration the specific user's triggers. Probability predictions were plausibly generated by GPT-5. |
| Anticip8 | Anticip8 | Anticip8 successfully considered the user's clinical context, producing nuanced failure nodes based on Rosa's susceptibility to sundowning. Anticip8's probability recalculations based on interventions were often inconsistent. |
| GPT-5 | Anticip8 | This combination ignored the complex dementia profile when generating failure nodes. Anticip8 produced unpredictable probability predictions, failing to reflect the actual benefit of the applied interventions. |
| Anticip8 | GPT-5 | Anticip8 accurately reflected the user's environmental vulnerabilities, generating relevant nodes such as "Transit hub confusion" and "Sensory overload". GPT-5 made plausible predictions about effects of interventions on probabilities. |

# 6 Field Testing

## 6.1 Methodology

Seven people volunteered to download the mobile application and use it for their daily journeys for a week. Since the users were not part of the target demographic, they were provided with SafeStep accounts set up with the two profiles given in Table 2. Users were prompted to read their profile description and keep it in mind while interacting with the application. The app was also instrumented to record data from the users. At the end of the testing period users were asked to provide answers, on a scale of 0-5, to the following questions:

1. Is the text size, color and contrast appropriate?
2. Are the visual elements in the app simple and not overwhelming?
3. Was the information displayed in the app useful and easy to understand?
4. Did the voice instructions accurately represent the text displayed?
5. Were the interventions reassuring or panic-inducing? 0=Highly panic inducing; 5=Highly reassuring.

The questionnaire also had a free-text field.

## 6.2 Results

Participants each used SafeStep 3-5 times on a total of 26 public transport journeys. The instrumentation metrics showed that there were 10 instances in which the users got off-track on their journeys. Users reported that when they got off track the system recognized this within 30 seconds.

In their responses to the questionnaire, users were mostly happy with the visual information displayed in the app, the voice instructions and the interventions (see Fig. 5). However, they gave low ratings for the font size, color and contrast. Given that the older demographic tends to struggle with font size, this should be taken into consideration when the application's interface is redesigned. In their free-text responses, most users reported interventions to be generally appropriate for the intended demographic, describing them as being "Very helpful" and "Properly worded and paced". One user mentioned that their interventions were not adequate for the stage they were in and emphasized the intervention's lack of relevance.

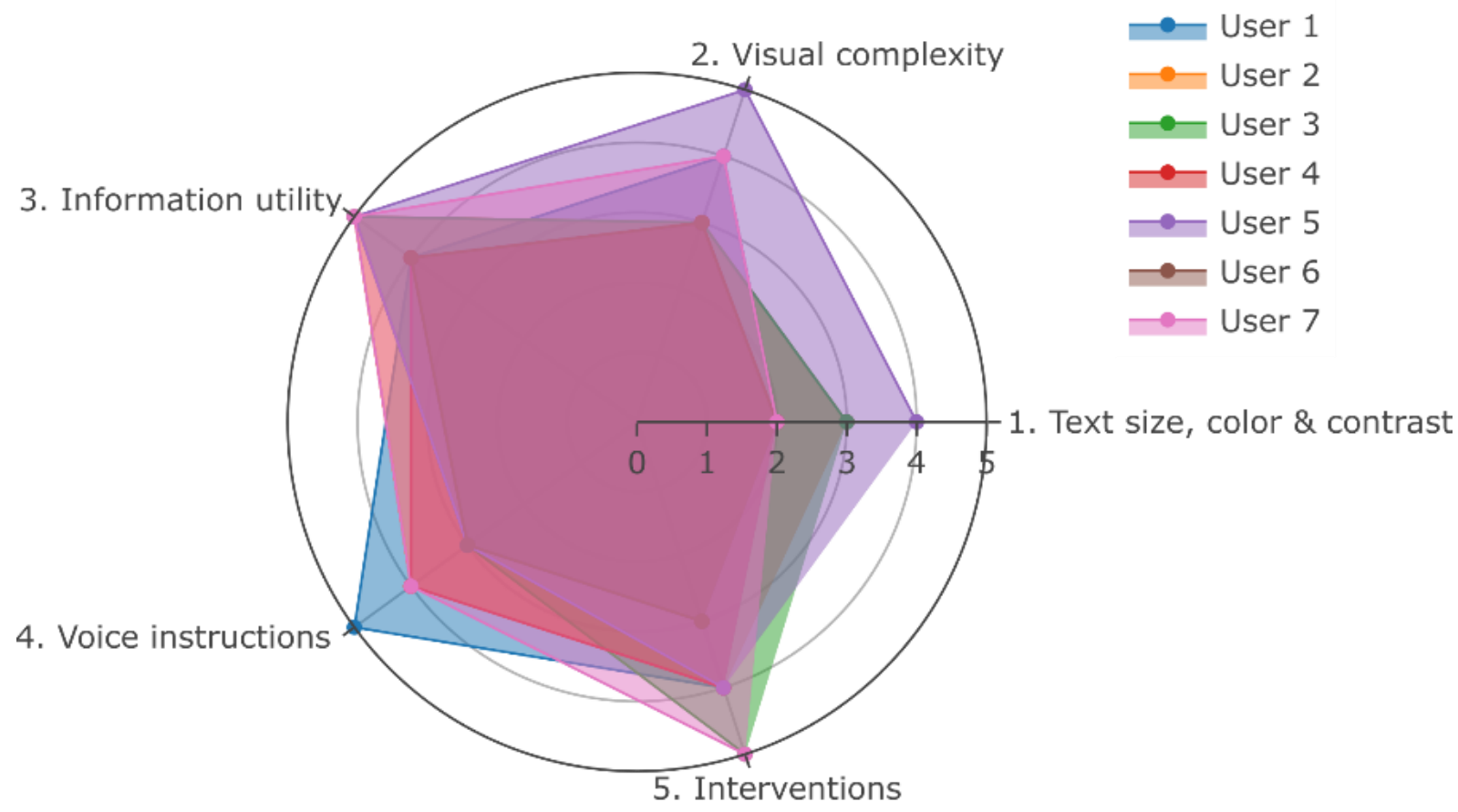

**Fig. 5.** Results from user questionnaire

## 7 Discussion

Feedback from test subjects strongly indicated that the app's interface needs to be simplified to improve usability for older people. The chat interface for journey planning could be replaced by a personal assistant avatar, and many other features could be improved. The font size, color and contrast should also be altered to make it more readable for older users.

The travel graph generation could be improved by storing a history of the failure nodes that occurred for each user and the interventions that were deployed when no failure occurred. This could be shared anonymously between users (with their explicit consent). This historical data could be included in the prompts used by SafeStep to generate the travel graph, which would enable it to provide more personalized failure nodes and suggest interventions that worked for users in similar situations in the past. Data about previous interventions and failures could also help to improve the probability predictions, shifting from the somewhat ad hoc numbers generated by GPT-5 to proper statistics derived from user data. To further improve performance, GPT-5 and Anticip8 could be fine-tuned with anonymized user data.

The travel graph generation process could also be improved if it had access to additional data sources. These could include weather information, visibility, traffic, event information (marches, football games, etc.), roadworks, and physiological data from a smart watch or fitness tracker (to monitor users' stress levels). Additional data would make it easier for SafeStep to respond to the real conditions that users encounter as they travel through a city.

In the current version of SafeStep the interventions are limited to text and voice. In many cases, this is not enough to steer the user away from a failure scenario. A future version of SafeStep could make more interventions available to the system, such as calling a taxi or carer, and SafeStep could connect with support mechanisms that are already available on public transport, such as the Passenger Assist service provided by Transport for London. For performance reasons, the interventions were tested individually, and a single intervention was selected for each stage of the journey. In the future, different combinations of interventions could be tried - the combination that was predicted to perform best would be selected for the user.

Unfortunately, time, budget and ethical constraints prevented us from testing SafeStep on elderly users. In the future, SafeStep should be tested on the target demographic, possibly in collaboration with a charity that works with the elderly. This testing would have to be carried out with a carer present to reduce the chances that the app puts an elderly person at risk of harm.

The SafeStep AI graph planner could be used in many other areas. For example, it could help people who are recovering from drug addiction, guide people who are working on their mental health, and assist with career coaching. In each application, a language model could be used to generate the stages of the person's journey (to recovery, mental health or career success), and then GPT-5 or Anticip8 could propose failure nodes and test interventions for each stage of the journey.

## 8 Conclusion

This paper has described an early prototype of a system that is designed to help people with frailty and dementia to travel in urban environments. Once the journey is planned, a combination of Anticip8 and GPT-5 is used to predict how the journey could go wrong and test interventions that could prevent the failure scenarios. The system consists of a travel graph generator and an app, which guides the user on their journey and deploys the best interventions to keep the journey on track.

Experiments on the travel graph generation indicated that Anticip8 is well suited to generating failure scenarios and GPT-5 is more capable at predicting how interventions affect the probability of failure scenarios. User feedback from field testing was generally positive, although there is clearly work to do on the interface.

Future plans for SafeStep include improvements to the interface and travel graph generation, and it needs to be fully tested on its target users. It can then be made available to assist people with frailty and dementia on their journeys. This will help them to engage with society, perform activities that support their physical and mental health, and potentially reduce their need for permanent care in a home. The AI graph planner that was developed for SafeStep could also be used in other areas, such as mental health, career coaching and recovery from addiction.

**Disclosure of Interests.** The work described in this paper was a student project that was partially funded by Fountech AI Limited. Fountech is the developer and owner of Anticip8.

## References

1. Office for National Statistics, https://www.ons.gov.uk/peoplepopulationandcommunity/populationandmigration/populationestimates/bulletins/annualmidyearpopulationestimates/mid2024, last accessed 2026/2/10
2. British Geriatrics Society https://www.bgs.org.uk/sites/default/files/content/attachment/2025-02-25/BGS%20Frailty%20Briefing%20-%20Feb%202025.pdf, last accessed 2026/2/14
3. Jung, D.: Fear of Falling in Older Adults: Comprehensive Review. Asian Nursing Research **2**(4), 214–222 (2008)
4. Alzheimer's Society, https://www.alzheimers.org.uk/blog/how-many-people-have-dementia-uk, last accessed 2026/2/16
5. Alzheimer's Society https://www.alzheimers.org.uk/about-dementia/stages-and-symptoms/anxiety-dementia last accessed 2026/2/15
6. McShane, R., Gedling, K., Keene, J., Fairburn, C., Jacoby, R., Hope, T.: Getting Lost in Dementia: A Longitudinal Study of a Behavioral Symptom. International Psychogeriatrics **10**(3), 253–260 (1998)
7. Roberts, J., Roberts, L., Reed, A.: Supporting the Digital Autonomy of Elders Through LLM Assistance. Proceedings of the AAAI Symposium Series **4**(1), 182–186 (2024)
8. Gong, L., Lin, Y., Zhang, X., Lu, Y., Han, X., Liu, Y., Guo, S., Lin, Y., Wan, H.: Mobility-llm: Learning visiting intentions and travel preference from human mobility data with large language models. Advances in Neural Information Processing Systems **37**, 36185-36217 (2024)
9. Tahir, H., Özdemir, M.K., Alhajj, R.: The role of LLM-powered chatbots in assisting elderly people: systematic review. Network Modeling Analysis in Health Informatics and Bioinformatics **15**(1), 26 (2026)
10. Kong, Y., Fu, Y.: Human Action Recognition and Prediction: A Survey. International Journal of Computer Vision **130**, 1366–1401 (2022)
11. Kolekar, S., Gite, S., Pradhan, B., Kotecha, K.: Behavior Prediction of Traffic Actors for Intelligent Vehicle Using Artificial Intelligence Techniques: A Review. IEEE Access **9**, 135034-135058 (2021)
12. Rao, S.P., Cook, D.J.: Predicting Inhabitant Action using Action and Task Models with Application to Smart Homes. International Journal of Artificial Intelligence Tools **13**(1), 81-99 (2004)
13. Auletta, F., Kallen, R.W., Bernardo, M.d., Richardson, M.J.: Predicting and understanding human action decisions during skillful joint-action using supervised machine learning and explainable-AI. Scientific Reports **13**, 4992 (2023)

14. Dakalbab, F., Talib, M.A., Waraga, O.A., Nassif, A.B., Abbas, S., Nasir, Q.: Artificial intelligence & crime prediction: A systematic literature review. Social Sciences & Humanities Open **6**(1), 100342 (2022)
15. Takeyama, K., Liu, Y., Sra, M.: TR-LLM: Integrating Trajectory Data for Scene-Aware LLM-Based Human Action Prediction. Proceedings of the IEEE/RSJ International Conference on Intelligent Robots and Systems (IROS), 18708-18715 (2025)
16. Graule, M.A., Isler, V.: GG-LLM: Geometrically Grounding Large Language Models for Zero-shot Human Activity Forecasting in Human-Aware Task Planning. Proceedings of the IEEE International Conference on Robotics and Automation (ICRA), 568-574 (2024)
17. Zhao, Q., Wang, S., Zhang, C., Fu, C., Do, M.Q., Agarwal, N., Lee, K., Sun, C.: Antgpt: Can large language models help long-term action anticipation from videos? International Conference on Learning Representations, 56677-56697 (2024)
18. Ogawa, T., Yoshioka, K., Fukuda, K., Morita, T.: Prediction of actions and places by the time series recognition from images with Multimodal LLM. Proceedings of the IEEE 18th International Conference on Semantic Computing (ICSC), 294-300 (2024)
19. Song, Y., Li, J., Bian, Y., Cai, Z.: Predicting User Behavior in Smart Spaces with LLM-Enhanced Logs and Personalized Prompts. Proceedings of the AAAI Conference on Artificial Intelligence **39**(1), 764–772 (2025)
20. Kolluri, A., Wu, S., Park, J.S., Bernstein, M.S.: Finetuning LLMs for Human Behavior Prediction in Social Science Experiments. Proceedings of the 2025 Conference on Empirical Methods in Natural Language Processing, 30096–30111 (2025)
21. Debenedetti, A.:SafeStep - The Proactive Journey Companion. Bachelor's thesis, Middlesex University (2026). Available at: https://github.com/KitchenTile/COMPANION_APP_BACKEND/